\pdfoutput=1
\documentclass[preprint, 10pt]{elsarticle}
\usepackage{amssymb}
\usepackage{amsthm}
\usepackage{amsfonts}
\usepackage{tikz}
\usepackage{colortbl}
\usepackage{comment}
\usepackage{multirow}
\usepackage{url}
\usepackage{algorithm}
\usepackage{algorithmicx}
\usepackage{textcomp}
\usepackage{bm}
\usepackage{subfig}
\usepackage{stfloats}
\usepackage{tabularx} 
\usepackage{diagbox}
\usepackage{xcolor}
\graphicspath{figures}
\DeclareGraphicsExtensions{.pdf,.jpeg,.png}
\definecolor{gray}{rgb}{0.85,0.85,0.85}
\newcolumntype{P}[1]{>{\centering\arraybackslash}p{#1}}
\newcolumntype{?}{!{\vrule width 1pt}}
\usepackage[export]{adjustbox}
\usepackage{hyperref}

\makeatletter % changes the catcode of @ to 11
\def\ps@pprintTitle{%
    \let\@oddhead\@empty
    \let\@evenhead\@empty
    \def\@oddfoot{\tiny\itshape
      {Preprint accepted at Digital Signal Processing, \href{https://creativecommons.org/licenses/by-nc-nd/4.0/}{CC-BY-NC-ND 4.0 license.}} \hfill November 1, 2020 |  \url{https://doi.org/10.1016/j.dsp.2020.102907}
}%
    \let\@evenfoot\@oddfoot
    }

\makeatother % changes the catcode of @ back to 12

\begin{document}

\begin{frontmatter}



\title{Optimized Deep Encoder-Decoder Methods for Crack Segmentation}


\author[label1]{Jacob K\"onig\corref{cor1}}
\ead{jacob.konig2@gcu.ac.uk}
\author[label1]{Mark David Jenkins}
\author[label1]{Mike Mannion}
\author[label1]{Peter Barrie}
\author[label1]{Gordon Morison}

\cortext[cor1]{Corresponding Author}

\address[label1]{School of Computing, Engineering and Built Environment, Glasgow Caledonian University, G4 0BA Glasgow, Scotland}

\begin{abstract}
Surface crack segmentation poses a challenging computer vision task as background, shape, colour and size of cracks vary. 
In this work we propose optimized deep encoder-decoder methods consisting of a combination of techniques which yield an increase in crack segmentation performance. Specifically we propose a decoder-part for an encoder-decoder based deep learning architecture for semantic segmentation and study its components to achieve increased performance. We also examine the use of different encoder strategies and introduce a data augmentation policy to increase the amount of available training data. The performance evaluation of our method is carried out on four publicly available crack segmentation datasets.
Additionally, we introduce two techniques into the field of surface crack segmentation, previously not used there: Generating results using test-time-augmentation and performing a statistical result analysis over multiple training runs. The former approach generally yields increased performance results, whereas the latter allows for more reproducible and better representability of a methods results.
Using those aforementioned strategies with our proposed encoder-decoder architecture we are able to achieve new state of the art results in all datasets.
\end{abstract}

\begin{comment}
\begin{graphicalabstract}
\end{graphicalabstract}

\begin{highlights}

\end{highlights}
\end{comment}

\begin{keyword}
Crack Segmentation \sep Convolutional Neural Network \sep Deep Learning \sep Semantic Segmentation



\end{keyword}

\end{frontmatter}



\section{Introduction}

Regular usage, ageing and environmental conditions are some of the main factors which contribute to the deterioration of road surfaces. In combination with the ever expanding urbanization and development of public infrastructure, the surface area of roads grows and so does the need for maintenance to keep those structures in usable condition. One of the main defects that can appear on road surfaces are cracks \cite{salman2013PavementCrack, eisenbach2017HowGet}. If not treated within a reasonable timeframe these cracks can increase in size, impair the structural integrity of roads and pavements, facilitate further defects such as potholes as well as lead to material damage and required maintenance downtimes \cite{eisenbach2017HowGet, zou2019DeepCrackLearning}. Therefore it is important to regularly survey those surfaces and fix faults before they escalate.
Due to the developments in information technology and image processing techniques  it is now possible to enable the partly automation of the surface inspection process to detect faults such as cracks. Proposed systems can include a variety of sensors such as cameras or laser \cite{eisenbach2017HowGet, yamaki2017RoadDeformation, fan2019RealTimeDense} as well different platforms on which those sensors are deployed. These may be groundborne, such as specialised vehicles \cite{eisenbach2017HowGet} or airborne as in UAVs \cite{fan2019RealTimeDense}. However, independent of the platform or the sensors used, a large amount of data is collected which needs to be analysed, either manually, which is unduly time consuming, or by the use of automatic algorithmic methods.

\begin{figure}[h]
\centering
\includegraphics[width=0.5\linewidth]{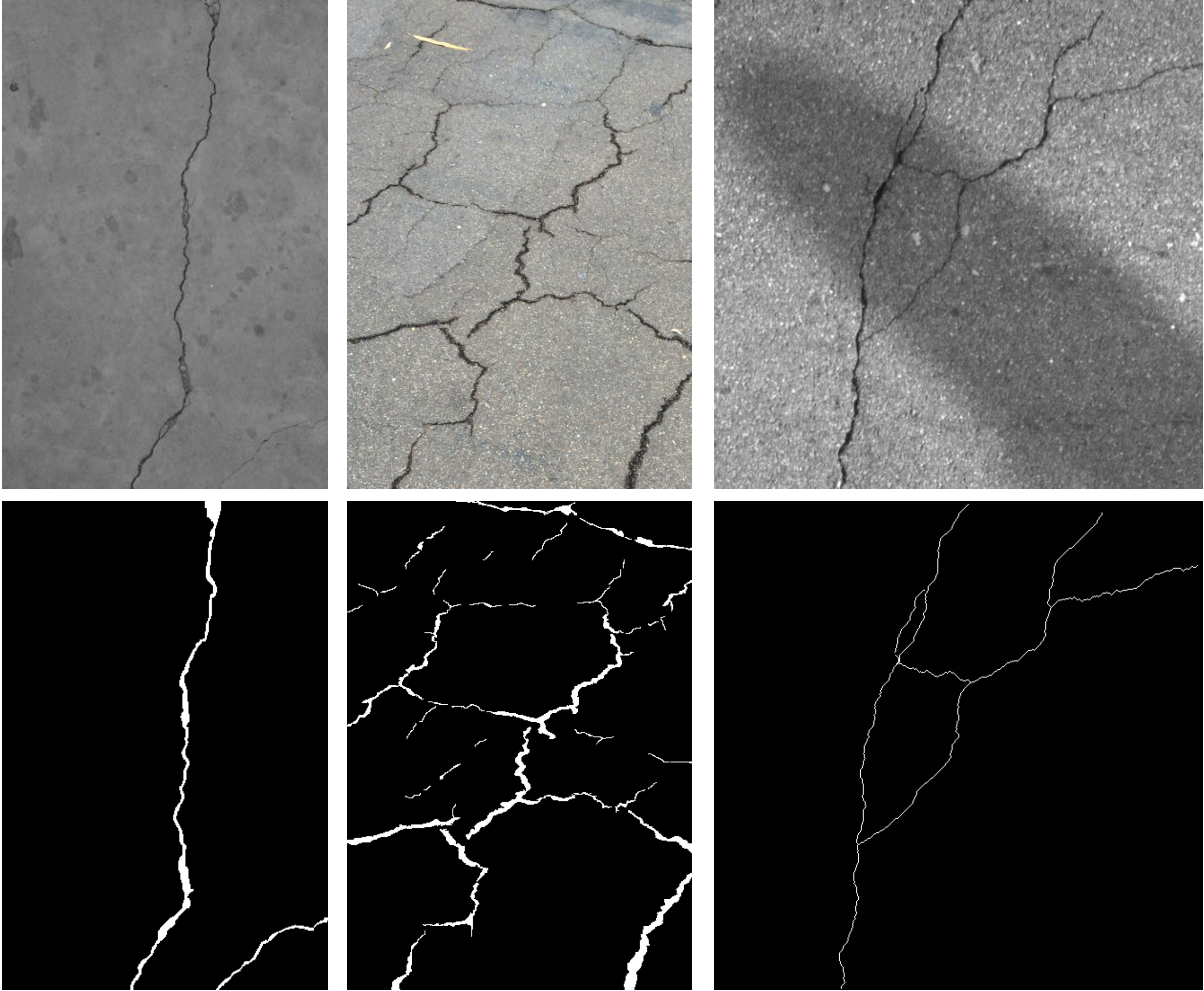}
\caption{Sample crack images (top row) and their annotated ground truth labels (bottom row) from three datasets (\cite{shi2016AutomaticRoad,liu2019DeepCrackDeep,zou2019DeepCrackLearning}), showing some of the challenges of the crack segmentation task: non-uniform backgrounds, obstructions such as shadows, dirt and stains, differing crack widths as well as changing camera angles. }
\label{fig_cracksamples}
\end{figure}

\figurename \ \ref{fig_cracksamples} shows the diversity of surface cracks, they may traverse into the horizontal and vertical directions, appear in grid like patterns, vary in width along the same crack and appear atop of different surface structures. In addition to that, other anomalies may cover the appearance of cracks in images, including road markings, dirt patches, oil stains and vegetation, all of which are common occurrences. This makes this task increasingly more difficult when conducted manually, in addition to any subjectivity an annotator might have \cite{salman2013PavementCrack, oliveira2013AutomaticRoad}.

In the computer vision domain, deep learning based convolutional neural networks (CNN) provide state of the art performance in a variety of tasks such as classification, detection and segmentation \cite{lin2019RefineNetMultiPath,tan2019efficientnet, tan2019EfficientDetScalable}. In some tasks they have even surpassed human performance \cite{he2015DelvingDeep}. Other work has also shown that when working on images with cracks, deep learning based methods outperform other techniques using handcrafted features or further machine learning approaches, making them well suited for the analysis of cracks in images \cite{zou2019DeepCrackLearning, liu2019FPCNetFast, liu2019DeepCrackDeep, fan2019RealTimeDense, zhang2016RoadCrack}. In this work we follow the popular approach of using semantic segmentation to analyse surface images, automatically annotating an image on a pixel level and stating for each pixel if it either contains a crack or not. Other approaches have been proposed to classify whether images or patches contain cracks \cite{wang2017GridBasedPavement, zhang2016RoadCrack} or locate cracks through creation of bounding boxes \cite{carr2018RoadCrack}, however segmentation provides a better level of detail to assess the severity of cracks and allows for tracking of potential crack changes over time.

Encoder-decoder based CNN architectures have achieved significant results, from segmentation of city-scenes \cite{badrinarayanan2017SegnetDeep, chen2018EncoderDecoderAtrous}, to medical image segmentation \cite{ronneberger2015UNetConvolutional} as well as for crack segmentation \cite{zou2019DeepCrackLearning, liu2019FPCNetFast, yang2019FeaturePyramid, konig2019ConvolutionalNeural}. This is a popular approach as the encoder learns denser features and the decoder then increases the size of the features again for creation of a segmented output.
Connecting features of the encoder and decoder parts in an architecture, not only at the bottleneck such as in SegNet\cite{badrinarayanan2017SegnetDeep}, but at their interim stages generally leads to improved results.
This is shown in state of the art results in the crack domain, such as in DeepCrack \cite{zou2019DeepCrackLearning}, which fuses multi-scale features at the different levels of the encoder-decoder architecture, FPCNet \cite{liu2019FPCNetFast} which sums skipped feature maps in the decoder upsampling process or as in the architecture in \cite{konig2019ConvolutionalNeural}, using concatenated shortcut connections between encoder and decoder as introduced in the U-Net \cite{ronneberger2015UNetConvolutional} architecture for medical image segmentation.

Several recent encoder-decoder segmentation architectures have been designed around using an encoder backbone which has been derived from popular image classification architectures \cite{liu2019DeepCrackDeep, song2020AutomatedPavement, chen2018DeepLabSemantic, qian2019AutomatedDetection}. As these encoders are designed for image classification they provide strong feature extraction capabilities and are therefore ideal for use as an encoder part in segmentation architectures.

Based on those observations we present an universal decoder design which can be added to a variety of popular feature-encoding backbones such as VGG \cite{simonyan2015VeryDeep}, ResNet \cite{he2016DeepResidual} and EfficientNet \cite{tan2019efficientnet} to effectively segment surface cracks.

Whilst many of the previous works only report their results on a small amount of datasets \cite{liu2019FPCNetFast, konig2019ConvolutionalNeural, inoue2019DeploymentConscious, liu2019DeepCrackDeep}, we show that our decoder-design in conjunction with a pretrained backbone achieves high performance across a multitude of datasets.

In summary, our proposed optimized deep encoder-decoder methods consist of the following contributions:
\begin{enumerate}
    \item We design a novel decoder part for encoder-decoder based CNN architectures to segment surface cracks in images. This decoder can be added to different encoder architectures, completing an U-Net like shape for the task of semantic segmentation. The components of this decoder architecture have been selected to optimize the performance on the task of crack segmentation.
    \item The performance of this decoder is then evaluated on five relevant crack datasets. It is shown that using this decoder, in conjunction with an \textit{EfficientNet B5} encoder backbone, new state of the art results are achieved in all datasets.  To the best of our knowledge this is one of the first comparisons across a such a large number of crack segmentation datasets from different works.
    \item During evaluation, we apply two techniques which previously have not been used in the crack-segmentation domain; Firstly, we discover that test-time-augmentation, through resizing of images, yields an additional boost in performance and secondly, reporting of results using a single training run is error-prone and does not accurately present the performance of a model. We therefore propose to report results after performing a statistical analysis and do so by showing the performance of our model by reporting the average and standard deviation over ten training runs

\end{enumerate}
The reminder of this paper is organized as follows: related work in regards to crack segmentation is reviewed in Section \ref{sec:litreview}. In Section \ref{sec:architecture} we outline the details of our proposed encoder-decoder architecture and state the training configurations used. Section \ref{sec:experiments} describes the datasets and the experiments which have been carried out. This Section also reports the results and contains the experiments which have been carried out to justify the components of our architecture as well as our methods generalization abilities and limitations. This work is then concluded in Section \ref{sec:conclusion}.

\section{Crack Segmentation Background}
\label{sec:litreview}
Making use of the growing capabilities in image capturing and processing, automatic crack detection and segmentation approaches have widely been studied in recent years \cite{mohan2018CrackDetection}.

\subsection{Traditional Methods}

Early approaches in crack segmentation exploited photometric- and geometrical characteristics of cracks, e.g. differing colors and shapes of cracks compared to their background \cite{chambon2011AutomaticRoad}.
These include crack segmentation through thresholding and mathematical morphology \cite{tanaka1998CrackDetection, maode2007PavementCrack, peng2015ResearchCrack,  fujita2006MethodCrack, oliveira2009AutomaticRoad}. Further approaches make use of different filtering techniques like using Wavelet transform \cite{subirats2006AutomationPavement, wang2007WaveletBasedPavement, zhou2005WaveletBasedPavement}, Gabor filters for segmentation \cite{salman2013PavementCrack} as well as edge detection algorithms who have also been exploited for segmentation of cracks such as in \cite{zhao2010ImprovementCanny} where the Canny edge detector \cite{canny1986ComputationalApproach} is used.  

However, these early approaches are limited in their performance due to the varying appearance of cracks and the amount of tuning required to make them work on various datasets. Hence, machine learning methods have started to gain traction, providing the ability to learn based on their input features and increase the performance over those previous methods \cite{makantasis2015DeepConvolutional, shi2016AutomaticRoad}. 

In \cite{shi2016AutomaticRoad}, Random Structured Forests are used to segment image patches and Support Vector Machines (SVM) or k-nearest neighbor (KNN) classifiers are used to classify whether this patch contains a crack. The work in \cite{fernandes2014PavementPathologies} uses a SVM to segment cracks based on previously extracted graph based features.
\subsection{Deep Neural Network Methods}

Accompanied by the general rise of using deep learning based methods for computer vision tasks, several works have also shown that using deep learning based techniques on crack images outperform those previous methods by very large margins \cite{zou2019DeepCrackLearning, zhang2016RoadCrack, jenkins2018DeepConvolutional, fan2018AutomaticPavement, inoue2019DeploymentConscious, makantasis2015DeepConvolutional}.

The work in \cite{makantasis2015DeepConvolutional} compares a multitude of traditional crack segmentation methods with a CNN architecture to segment tunnel wall defects. This work shows that using CNN to segment defect areas outperforms previous methods such as SVM and KNN as well as classification trees.

In \cite{fan2018AutomaticPavement} a CNN architecture is introduced which classifies whether the central pixel in an image patch contains a crack. Its results are then averaged for each pixel in an image to generate a segmentation map. The work in \cite{inoue2019DeploymentConscious} uses a similar approach with dilated receptive fields to create crack segmentation maps, in addition to using input augmentation and model-weight-sharing to predict the orientation of cracks. In \cite{fan2020EnsembleDeep} it is shown that an ensemble of CNN models can not only produce accurate segmentation maps but also provide crack-measurements. However, the drawback of this ensemble method is the lack of an end-to-end training approach.

Several recent deep learning architectures for crack segmentation are based on an encoder and decoder basis \cite{jenkins2018DeepConvolutional, carr2018RoadCrack, zou2019DeepCrackLearning, yang2019FeaturePyramid, liu2019FPCNetFast}. Inside the encoder, convolutional and downsampling operations create denser and spatially smaller feature maps which are then upsampled in the decoder to create an output segmentation maps, which matches the spatial dimensions of the input.
In \cite{jenkins2018DeepConvolutional, konig2019ConvolutionalNeural} architectures based on U-Net  \cite{ronneberger2015UNetConvolutional} are presented. U-Net encompasses shortcut connections, between the encoder and decoder part, with the goal of retaining spatial features lost due to pooling. The DeepCrack architecture in \cite{zou2019DeepCrackLearning} is based on SegNet \cite{badrinarayanan2017SegnetDeep}, but it contains a multi scale fusion component, extracting and making use of features from multiple scales of the feature pyramid to generate the segmentation map. In comparison with several other baseline CNN segmentation architectures such as U-Net, SegNet and hollistically-nested edge detection (HED) \cite{xie2015HolisticallyNestedEdge}, this architecture achieved superior results when trained on one, and evaluated on three custom datasets. 
FPCNet \cite{liu2019FPCNetFast} combines an encoder-decoder architecture with a module in the bottleneck which uses dilated convolutions to extract features of multiple scale in feature maps before using an adaptive upsampling approach which adds encoder features to the feature maps of the decoder and channelwise attention through the squeeze-and-exitation \cite{hu2017SqueezeandExcitationNetworks} method before further processing. The work in \cite{yang2019FeaturePyramid} expands upon the HED method \cite{xie2015HolisticallyNestedEdge} adding a feature pyramid module \cite{lin2017FeaturePyramid} as well as a component to adaptively reweight features of different levels of those feature pyramid before generating a segmentation result. Another architecture, also called DeepCrack, in \cite{liu2019DeepCrackDeep}, uses a VGG backbone \cite{simonyan2015VeryDeep} and a deeply supervised \cite{lee2015DeeplySupervisedNets} approach to uspcale and fuse the feature maps from all levels of the backbone, before applying guided filtering \cite{he2010GuidedImage} using the fused feature maps as well as a side output of the first convolution stage of the backbone, to create a segmentation output. The crack segmentation method in \cite{song2020AutomatedPavement} uses a ResNet backbone feature encoder and decodes the features maps through a multi-dilation module which uses dilated-convolutions to  extract crack-features from different scales. It also employs multi-scale fusion, merging interim sparse feature maps with dense ones from later layers.

\section{Crack Segmentation Architecture}
\label{sec:architecture}

\begin{figure}[h]
\centering
\includegraphics[width=\linewidth]{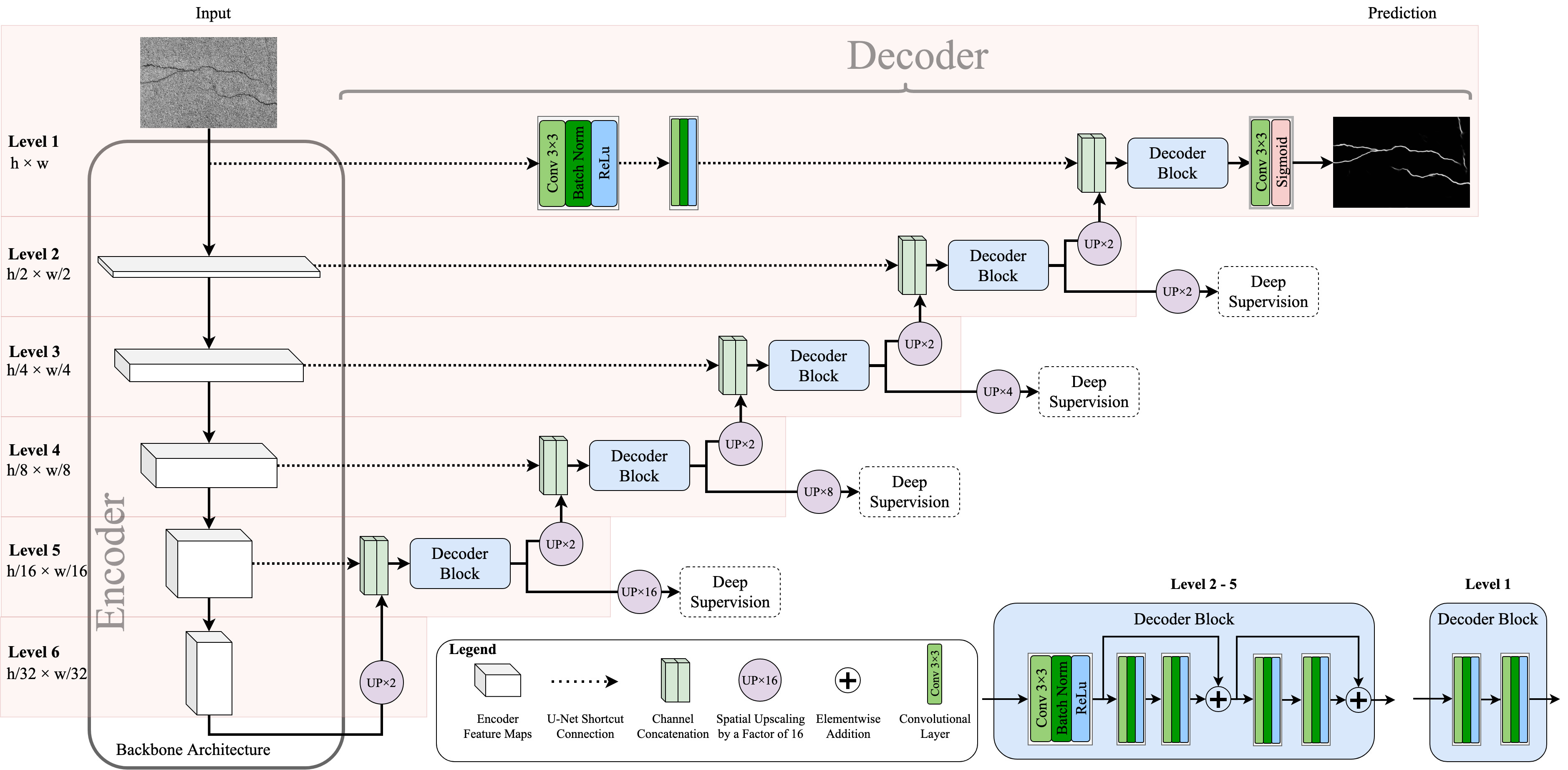}
\caption{Our proposed Crack Segmentation Architecture. Different encoder backbones can be used from which the U-Net shortcut connections are extracted by using the feature maps right before each downsampling operation. As illustrated, the decoder block in the first level differs from the remaining ones. Upsampling between levels is done through nearest-neighbor upsampling, whereas each upsampling operation before a deep supervision operation is chosen to be bilinear upsampling. The filters in the decoder are fixed as: [16, 32, 64, 128, 256], increasing with the level size.}
\label{fig_architecture}
\end{figure}

The following section introduces our architecture which is built following an U-Net \cite{ronneberger2015UNetConvolutional} like, encoder-decoder, shape and can make use of different pretrained feature encoders. To those encoders we add a decoder part, optimized for crack segmentation performance. An illustration of our architecture is shown in \figurename \ \ref{fig_architecture}.

\subsection{U-Net Basis}
U-Net based architectures contain an encoder-decoder like structure with skip-connections. Generally an input is propagated through successive convolutional, activation and pooling layers, reducing its feature size up to a bottleneck, therefore forming the encoder. In the following decoder, instead of pooling layers, upsampling layers are used to increase the feature size back to the spatial dimensions of the input. These sequences of using convolutions, activations and pooling allows the network to extract more condensed features, with the drawback of reducing the spatial size of the feature map. To counteract the loss of spatial information after this encoder part, U-Net therefore introduces skip-connections. These connections combine interim features of the encoder part  with the output of the upsampling operation of the same spatial size through channel-concatenation, hence reintroducing spatial features. This is followed by feeding those concatenated features into the decoder layer-sequences.

\subsection{Decoder Design}

Our proposed decoder can be added to a variety of encoder backbones. However, encoders which aim to be used with our proposed decoder, need to have six levels, each distinguishable by their different spatial size. These are addressed with level numbers $l{=}1$ to $l{=}6$, each having half the spatial size of its previous one. Hence, an encoder needs to have five downsampling layers.

For each but the last level of the encoder, we add a decoder block.  The decoder blocks are thus situated in levels $l{=}1$ to $l{=}5$ and are connected to the encoder through U-Net style skip connections. To increase the spatial size inbetween different levels of the decoder, we make use of nearest neighbor upsampling. 
The input into the decoder sequence is the upsampled output of the sixth level of the encoder backbone. 
After each upsampling operation, the upsampled feature maps are concatenated with the output of the encoder at the specific level before being propagated to the decoder block. 

Some backbones, such as ones based on EfficientNet or ResNet, apply a downsampling convolution right after their input layer. This means that the resulting output at level $l{=}1$ of the encoder equals that of the input image.

To therefore still extract features at this spatial size, we expand the skip connection at that first level, to include two $3\times3$ convolutional layers (Conv), each followed by Batch Normalization (BN) and a ReLU activation on an input.

The decoder blocks in level $l{=}2$ to $l{=}5$ are residual blocks, meaning they contain residual connections inbetween convolutions. Adding those residual connections for semantic segmentation tasks in convolutional blocks has shown to improve the performance in U-Net based architectures \cite{zhang2018road, alom2018RecurrentResidual}. Every residual block applies two $3\times3$ Conv-BN-ReLU sequences on an input. This input is then subsequently added to the output of the Conv-BN-ReLU sequence, thus creating a residual connection. Each decoder block in those levels contains a standard Conv-BN-ReLU sequence followed by two residual blocks. 
The number of filters used in all decoder blocks is based on its level $l$ in the architecture, leading to the following number of filters per layer: [$16_{l=1}, 32_{l=2}$, $64_{l=3}$,$128_{l=4}$,$256_{l=5}$].
The decoder block at level $l=1$ applies two Conv-BN-ReLU sequences. A final segmentation output is then generated through running the resulting feature maps of the decoder block at level $l{=}1$ through a $3\times3$ convolutional layer with 1 filter and a Sigmoid activation function.

To enforce the learning of more robust feature this architecture also deeply supervises \cite{lee2015DeeplySupervisedNets} the outputs of the decoder blocks on levels $l{=}2$ to $l{=}5$. This is achieved through a $1\times1$ convolution with one filter, followed by a Sigmoid activation and bilinear upsampling to the original spatial size.

\subsection{Encoder-Backbone Architectures}

Commonly, the encoders in encoder-decoder structures are either tailored and designed for the specific image segmentation task \cite{liu2019FPCNetFast, ronneberger2015UNetConvolutional} or use the same architectures to ones that have been used for image classification tasks without the fully connected components \cite{badrinarayanan2017SegnetDeep, liu2019DeepCrackDeep, qian2019AutomatedDetection}. In this work we focus on the latter, by considering three different types of backbone architectures: VGG,  ResNet and EfficientNet. The weights, having been pretrained on ImageNet, provide the encoder with a strong feature extraction baseline and enable a faster training convergence. For all backbones, we omit the fully connected components and use the output of the last operation before the classification part as the output. 

The VGG architecture contains basic successions of convolutional and activation layers, with max pooling layers used to reduce the spatial feature sizes. There are several variants such as\textit{ VGG 16}, with 16- , or \textit{VGG 19} with 19-weight layers. 
It has successfully been used as an encoder backbone for general segmentation architectures such as SegNet as well as for crack segmentation in the DeepCrack architectures in \cite{liu2019DeepCrackDeep, zou2019DeepCrackLearning}.

ResNet is an architecture design which makes use of skip-connections, allowing for networks with more layers as well as better gradient flow. Several sizes of those networks have been introduced ranging from variants with 26 layers (\textit{ResNet 26}) up to variants with over 1000 layers (\textit{ResNet 1001}). Its variants are also commonly used as encoder-backbones for segmentation \cite{chen2018DeepLabSemantic, kampffmeyer2018ConnNetLongRange, song2020AutomatedPavement}.

The EfficientNet architecture  introduced a new way of designing CNNs. It is proposed that instead of scaling the depth, width and image-resolution separately, they should be scaled equally. Based upon this proposal, a baseline architecture (\textit{EfficientNet B0}) which was found using neural architecture search \cite{tan2019MnasnetPlatformAware}, is scaled up seven times (\textit{EfficientNet B1 to \textit{EfficientNet B7})}. The consequently resulting scaled up architectures perform much better and faster on ImageNet \cite{russakovsky2015ImageNetLarge} and other datasets such as CIFAR \cite{krizhevsky2009LearningMultiple} whilst using much less parameters in comparison to other competitive approaches. EfficientNet based architectures have also been previously used for segmentation, such as in \cite{qian2019AutomatedDetection}.

Our results are reported using an \textit{EfficientNet B5} backbone. In later experiments we validate the performance of this and other backbones and justify our choice.

\subsection{Training Configuration}
Our proposed architecture is trained on patches of size $288 \times 288$ pixels similar to the implementation in \cite{liu2019FPCNetFast}. Compared to other segmentation domains, crack segmentation can be seen as a binary classification problem, as pixels can only be assigned a probability of containing a crack or not. Let $X$ be a training dataset consisting of $N$ samples $x$, $X = \{x_1, x_2 \ldots x_N\}$ as well as $Y$ being the respective ground truth dataset to $X$ consisting of $N$ samples $y$, $Y = \{y_1, y_2 \ldots y_N\}$ then each ground truth pixel $i$ of any given sample $y$ is $y_{i} \in [0,1]$.

Recall that this architecture contains a total of five outputs, four interim deeply supervised outputs at level $l=2$ to $l=5$ and the final segmentation output. During the training process, the sum of the binary cross entropy and the Dice loss is applied on each output and the total sum of the losses of all outputs is used as the training loss. The cross entropy loss can be used for semantic segmentation, however it may  be affected when the classes are highly unbalanced such as in the case of crack- to no-crack pixels, where a trained architecture might prefer to segment background pixels. A common approach to counteract this is issue is to either weight classes differently, or use a weighted cross entropy loss \cite{liu2019DeepCrackDeep, xie2015HolisticallyNestedEdge}. We however add the Dice loss \cite{milletari2016VnetFully}, which measures the overlap of the prediction and the ground truth so that the total loss takes into account both the segmented shape as well as smaller details. 
The loss $L$ for a prediction $\hat{y}$ consisting of $P$ pixels at a specific output of the network is therefore calculated as:

\begin{equation}
L= -\textstyle{\sum_{i}^{P}}y_{i}\big(\log{\hat{y}_{i}}-(1-y_{i})\log(1-\hat{y}_{i})\big) \\ + \frac{ 2\sum_{i}^{P} y_{i}\hat{y}_{i}+1}{ \sum_{i}^{P} y_{i} +  \sum_{i}^{P}\hat{y}_{i} +1}
\end{equation}

\section{Experiments and Results}
\label{sec:experiments}
This section describes the experiments which have been carried out on different datasets as well as further studies on how the different components of our method affect its performance.

\subsection{Datasets}
\label{sec:datasets}
This work makes use of five datasets, CrackForest \cite{shi2016AutomaticRoad}, DeepCrack-DB \cite{liu2019DeepCrackDeep}, CrackTree260, CRKWH100 and Stone331, in which the latter three were introduced in \cite{zou2019DeepCrackLearning}. 
\begin{itemize}
    \item The \textit{CrackForest} dataset\footnote{ Available at \url{https://github.com/cuilimeng/CrackForest-dataset}} (CFD) consists of a total of 118 images of size $480\times320$ pixels. The images have been taken from road surfaces in Beijing. We omit one image as it contains a wrong ground truth\footnote{The filename of the omitted image is \textit{042.jpg}}. This dataset is then split into 71 training, and 46 testing images\footnote{For reproducibility we use images \textit{001.jpg} to \textit{072.jpg} for training and the remainder for testing} similar to \cite{fan2018AutomaticPavement, inoue2019DeploymentConscious, liu2019FPCNetFast}. The images are supplied in RGB, however due to the small dataset size and little difference in color channels we convert them to grayscale.
    \item \textit{CrackTree260} (CT260) is a dataset containing 260 grayscale road pavement images of different sizes ($800\times600$ and $960\times720$ pixels). Similar to the approach in \cite{zou2019DeepCrackLearning} this dataset is exclusively used for training the architectures.
    \item The \textit{CRWKH100} dataset consists of 100 grayscale images road pavement images of size $512\times512$ pixels, all of which are used exclusively for testing.
    \item \textit{Stone331} consists of 331 grayscale images of stone surfaces, again of size $512\times512$ pixels. However, not the whole surface of the image contains the stone, therefore masks are provided to exclude predictions outside the stone area. This dataset is also used exclusively for testing.
    \item A total of 537 RGB images are contained in \textit{\mbox{Deepcrack-DB}}\footnote{Due to a naming overlap with the work in \cite{zou2019DeepCrackLearning} we refer to this dataset as \textit{Deepcrack-DB} as the authors have not given it a name}\cite{liu2019DeepCrackDeep}.The dataset split is given as 300 training-, and 237 testing images, all of size $544\times384$ pixels.  
\end{itemize}

In \cite{zou2019DeepCrackLearning} another dataset, CrackLS315 is used, however the authors were not able to provide us with this dataset. CT260, CRKWH100 and Stone331 can be seen as one large dataset, as algorithms trained on CT260 are tested on CRKWH100 and Stone331. It is observed that the CT260 and CRKWH100 images are of similar-looking nature, whereas Stone331 images appear different. This leads to generally worse performance in Stone331 compared to CRKWH100 (as our following experiments indicate) however it may also show the generalization performance of algorithms.

As it can be seen, the number of images in each dataset is limited compared to larger datasets such as ImageNet. Therefore we perform augmentation on the training images. At first we enhance the number of images by rotating each training image for $[0^{\circ}, 90^{\circ}]$ as well as mirroring it across the $[x, y, y{=}x, y{=}{-}x]$ axis. This means for each training image 8 variations of it are created, equaling the Dihedral group D4.

\begin{algorithm}[h]
    \small
    \caption{Detailed augmentation policy for the training process of our architecture.}
    \label{algo:aug}
    \textit{The bold character in brackets states if augmentation is applied to:
    \vspace{-5pt}
    \begin{itemize}
    \itemsep-.5em 
        \item only the image {\normalfont \textbf{(i)}}
        \item same augmentation to the image and mask {\normalfont \textbf{(i,m)}}
    \end{itemize}} 
    \vspace{-5pt}
    \textit{p states the probability with which each augmentation step is applied}
    \vspace{5pt}\\
    \textbf{FOR EACH training \textit{image} (i) and corresponding \textit{mask} (m) }:
    \begin{enumerate}
        \item Random crop of size $288\times288$ pixels \textbf{(i,m)}
        \item With $p{=}0.5$ apply  Random brightness shift in range $[-10\%,+10\%]$  \textbf{(i)}
        \item With $p{=}0.5$ apply Random contrast shift in range $[-10\%,+10\%]$ \textbf{(i)}
        \item Apply one of:
        \begin{enumerate}
            \item Random Additive Gaussian Noise with a standard deviation sampled in range of [0, 2.55] \textbf{(i)}
            \item Random Multiplicative Noise in range $[75\%,125\%]$ \textbf{(i)}
        \end{enumerate}
        \item With $p{=}0.3$ apply one of:
        \begin{enumerate}
            \item Random spatial reduction of the image in range $[0.75,1]$ with mirror padding back to $288\times288$ pixels \textbf{(i,m)}
            \item Random sized crop, minimum size of $144\times144$ pixels with nearest-neighbor upsampling back to $288\times288$ \textbf{(i,m)}
        \end{enumerate}
        \item \textbf{OUTPUT} \textbf{(i,m)}
    \end{enumerate}
\end{algorithm}

\begin{figure}[h]
\centering
\includegraphics[width=\linewidth]{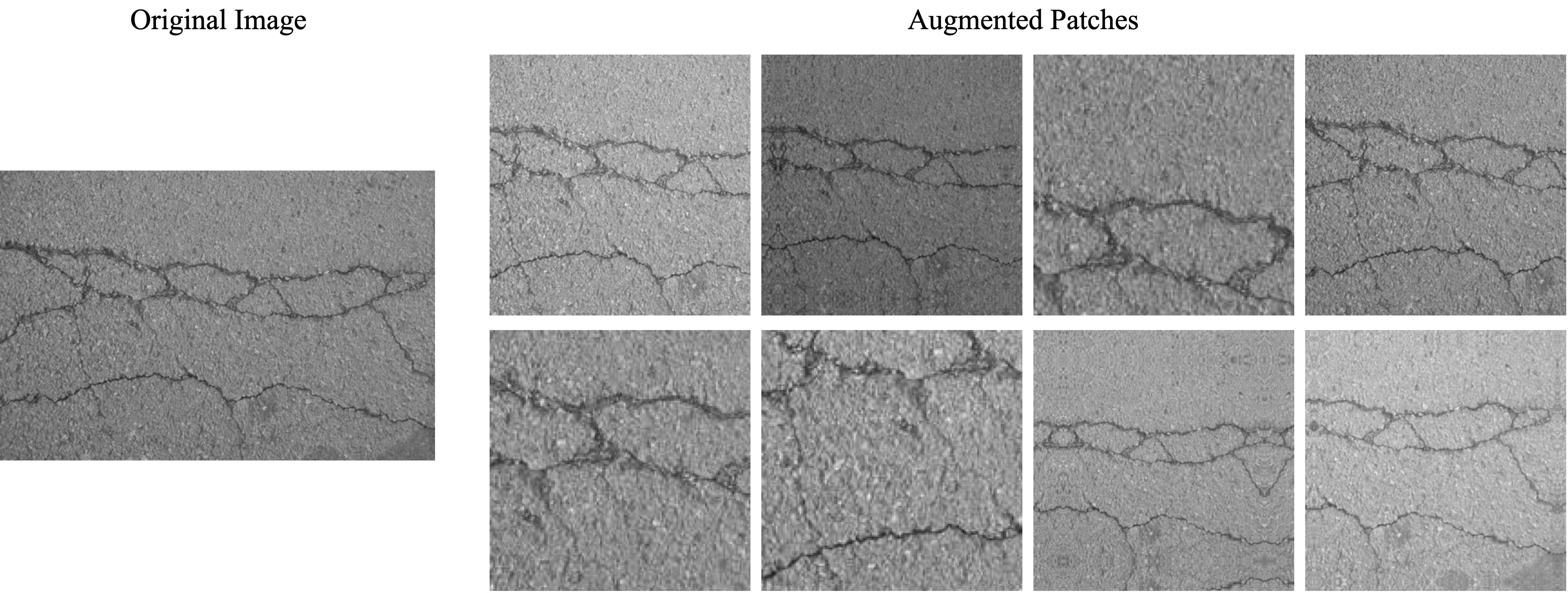}
\caption{Augmentation image samples when using an image of the CFD dataset. This augmentation policy creates quadratic image patches of $288\time288$ pixels size. The brightness and contrast changes emulate different backgrounds and crack-colourings, whereas the size-augmentation exposes the network to more crack data of different widths during training.}
\label{fig_augmentation}
\end{figure}

As previously mentioned our architecture is trained on images of size $288\times288$. This is achieved using an augmentation policy of random cropping, brightness and contrast shift, the addition of noise as well as random zooming in or zooming out, which results in an image of the selected size. Due to the fully convolutional nature of our architecture, as it does not contain any fully connected layers with a fixed number of connections which may restrict images sizes, evaluation can then be performed on the full sized testing images. The detailed augmentation policy is outlined in Algorithm \ref{algo:aug}. Sample images generated by our augmentation policy are shown in \figurename \ \ref{fig_augmentation}.

\subsection{Metrics}
In this work, results are reported using three metrics. All of them are based on  the harmonic mean: the $F1$ score of Precision $PR =  \frac{TP}{TP + FP}$ and Recall $RE = \frac{TP}{TP + FN}$. These can be calculated based on True Positive $TP$, False Negative $FN$ and False Positive $FP$ predictions. The $F1$ is calculated as $F1=2\cdot\frac{PR\cdot RE}{PR+RE}$. 
In the crack segmentation domain the \textit{F1} metric is applied three ways, all of which make use of an oracle. This oracle contains the solutions as well as all predictions with the aim to chose the best possible result for each metric.
\begin{enumerate}
    \item Reporting the aggregated $F1$ metric when an oracle extracts the best $F1$ score of all confidence thresholds of each image yields the \textit{OIS} (Optimal Image Scale) metric \cite{arbelaez2011ContourDetection}. 
    \item The \textit{ODS} (Optimal Dataset Scale) metric reports the aggregation of $F1$ scores from all images when an oracle has set the best possible fixed threshold across all images in the dataset \cite{arbelaez2011ContourDetection}.
    \item A cumulative ODS, \textit{cODS}, as used
    in \cite{jenkins2018DeepConvolutional, fan2018AutomaticPavement, liu2019FPCNetFast, liu2019DeepCrackDeep}, reports the $F1$ when $PR$ and $RE$ have been calculated across the whole dataset and not as an aggregation of results on each image. We assume that when no confidence threshold cutoff is reported in those works, the one that achieves the highest results extracted through an oracle has been used. Therefore we are referring to this metric as \textit{cumulative} ODS.
\end{enumerate}
The formulas for each metric are derived as follows:
\begin{equation}
    OIS =  \frac{1}{N_{img}}  \sum_{i}^{N_{img}}  max \Big\{ F1_t^i: \forall t \in \{0.01, ..., 0.99\} \Big\}
\end{equation}
\begin{equation}
    ODS = max \Big\{ \big\{ \frac{1}{N_{img}}  \sum_{i}^{N_{img}}   F1_t^i \big\} : \forall t \in \{0.01, ..., 0.99\} \Big\}
\end{equation}
\begin{equation}
    cODS = max \Big\{ F1_t^d : \forall t \in \{0.01, ..., 0.99\} \Big\}
\end{equation}
with $t$ denoting the confidence threshold at which a predicted pixel is counted as a $TP$, $N_{img}$ the number of images in the testing dataset $d$ and $i$ each specific image.

We only report the \textit{cODS} metric to be able to compare with other implementations. We propose to use the \textit{ODS} and \textit{OIS} metrics in further crack segmentation works, as \textit{cODS} does not accurately represent the real performance of an algorithm. Due to the way the $F1$ score is calculated there, it may skew results as a small proportion of images in a dataset may contain the majority of crack pixels.

As in previous works for crack segmentation \cite{jenkins2018DeepConvolutional, zou2019DeepCrackLearning, liu2019FPCNetFast, fan2020EnsembleDeep, konig2019ConvolutionalNeural, inoue2019DeploymentConscious}, in the experiments on the CFD, CT260, CRKWH100 and Stone331 datasets, $TP$ pixels are accounted for if they lie within two pixels of the corresponding ground truth, leading to a relaxed distance threshold. This distance threshold also changes the calculation of $FN$ and $FP$. A $FP$ pixel is counted if it is a wrong prediction farther away from a ground truth pixel than a two-pixel distance and a $FN$ pixel is counted as a missed prediction if the ground truth is farther away than two pixel from a prediction.

This was adopted as the labelling of cracks may be slightly inaccurate. The experiments on DeepCrack-DB do not use such two-pixel threshold to be able to provide an accurate comparison with \cite{liu2019DeepCrackDeep}.

\subsection{Experimental Configuration}
Our network architecture is trained using stochastic gradient descent as the optimizer using a degrading learning rate following the formula $lr_e=lr_{e=0}*0.96^{e}$ with $lr_e$ denoting the learning rate at a specific epoch $e$ and $lr_{e=0}$ denoting the initial learning rate at the first epoch $e{=}0$ with a value of 0.01. The weights in all layers are initialized using the method in \cite{he2015DelvingDeep}. The implementation of this work was achieved using the Keras-API \cite{chollet2015Keras} with the Tensorflow \cite{abadi2016TensorFlowSystem} backend and training and evaluation were performend on a NVIDIA Titan XP GPU. Parts of our architecture implementation make use of code from \cite{yakubovskiy2019EfficientNet}. On all training datasets the architecture is trained for 120 epoch, using a batch size of 8 and a weight decay of 1e-5. As the works introducing those datasets did not provide a subpart of the dataset for validation, we estimated the hyperparameters and training settings based on previous works \cite{liu2019FPCNetFast, zou2019DeepCrackLearning} and always used the weights of the last epoch for evaluation.
The results in the following section are reported without and with test-time-augmentation (TTA). Our results show that similar to the findings in \cite{li2017InstanceLevelSalient, kampffmeyer2018ConnNetLongRange} scaling the images to different sizes and aggregating the different predictions improves the segmentation results significantly. Due to the different image sizes, the CFD testing images are scaled in the factors [0.6, 0.8, 1.0, 1.2, 1.4] and in CRKWH100/Stone331 to  [0.5, 0.75, 1.0, 1.25, 1.5]. These scaling factors differ as the spatial dimensions of the image still need to be dividable by $2^5$, based on the five reductions of the spatial size of the feature maps by a factor of two in the encoder going from an input image of spatial size $h\times w$ to $\frac{h}{2^5} \times \frac{w}{2^5}$ in level 6 of the encoder.

On the DeepCrack-DB testing set we also use TTA, however due to the height-to-width ratio and the limitation of being dividable by $2^5$ we scale the images to the fixed sizes of [$288\times192$ , $416\times288$, $544\times384$, $672\times480$, $832\times576$]. Thereby each spatial dimension is scaled to the closest possible divisor of the range [0.5, 0.75, 1.0, 1.25, 1.5]. 

In addition to that we also found that repeatedly running each experiment, even with fixed data sampling seeds, do not yield consistent results due to non-deterministic GPU operations. This is similar to the findings in \cite{marrone2019ReproducibilityDeep}. Therefore we report our results as averages with their standard deviation over 10 runs. Across all different tested configurations, each experiment has the same fixed seed for selection and augmentation of training data, for each specific run. As we will see in the following sections, averaging those results is important as outliers may detract from the actual performance of those algorithms, therefore not representing the general model performance.

The approximate training run time of our proposed method using an \textit{EfficientNet B5} backbone for a single run with the aforementioned configuration is 1:20h for CFD, 4:50h for CT260 and 5:30h for DeepCrack-DB.

\begin{figure}[h]
\centering
\includegraphics[width=0.5\linewidth]{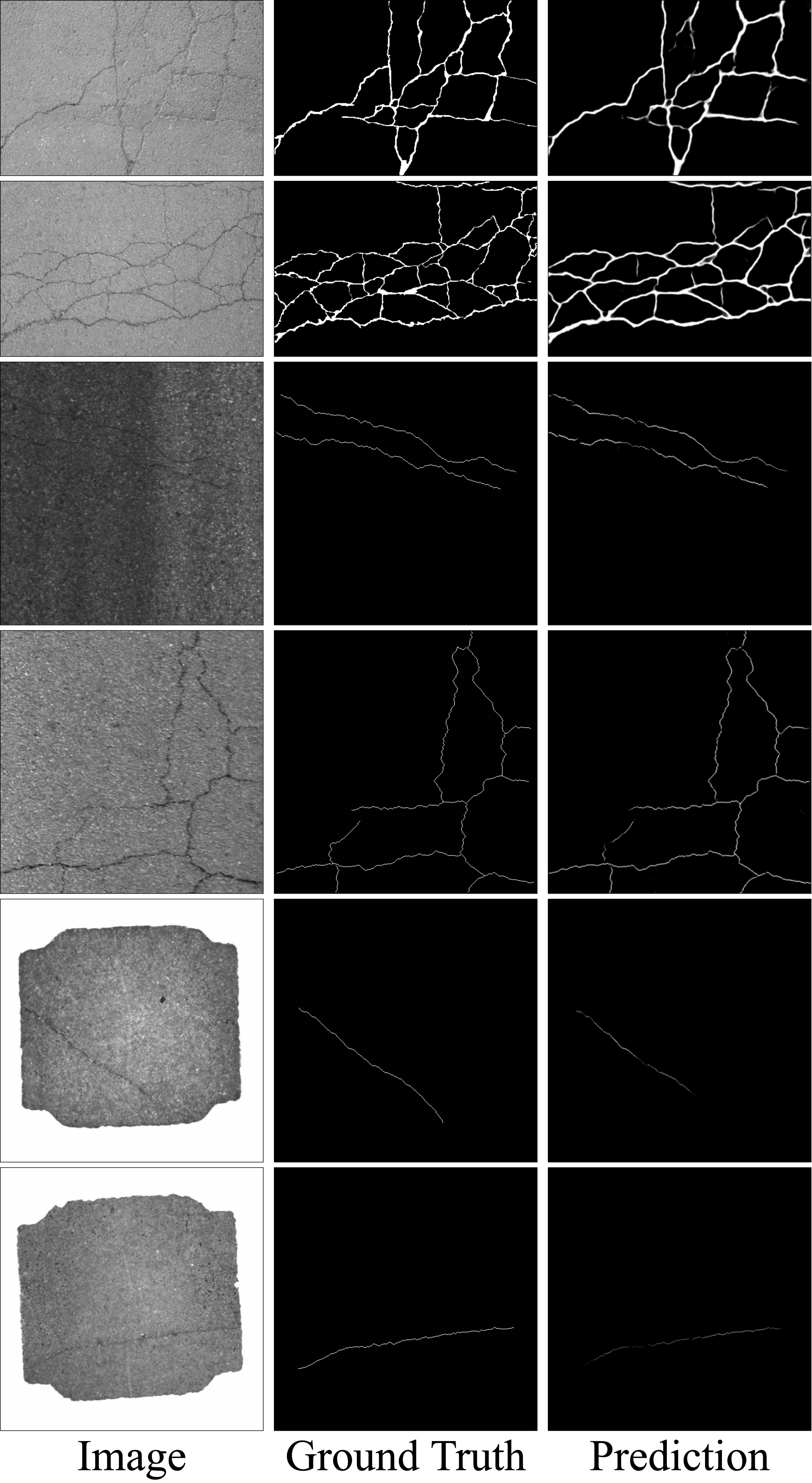}
\caption{Sample results of using our architecture on crack images from different datasets.}
\label{fig_results}
\end{figure}

Sample results of using our architecture are shown in \figurename \ \ref{fig_results}. We note that due to our reporting approach using averaged results and the results lying in range of ${>}85\%$ \textit{OIS}/\textit{ODS}, visible differences between our approaches and other implementations are very small, hence we chose to omit those visual comparisons and stick to reporting using the chosen metrics.

\subsection{Results}

\subsubsection{Results on CFD}
\label{sec:results_cfd}
The results on the CFD dataset are shown in \tablename \ \ref{table_cfd}. As it can be seen previous results report their performance in the \textit{cODS} metric. We show that our architecture achieves a slightly better average performance of 0.04\% when run without TTA. When testing the images on multiple scales we can report a total increase of 0.10\% in the \textit{cODS} metric. We also report the results for the \textit{OIS} and \textit{ODS} metrics, in which it also can be seen that TTA improves the results.

\begin{table}[htbp]
\centering
\footnotesize

\caption{Results on CFD. Our results are reported as (Avg)\textpm(Stdev) over 10 runs.}
\label{table_cfd}
\begin{tabular}{|l?r|r|r|}
\hline
\textbf{Method}  & $\boldsymbol{OIS}$ & $\boldsymbol{ODS}$& $\boldsymbol{cODS}$ \\
\hline
Fan \textit{et al.} \cite{fan2018AutomaticPavement} & - &  - & 92.44   \\
Fan \textit{et al.} (ensemble) \cite{fan2020EnsembleDeep} & - & - & 95.33 \\
Inoue and Nagayoshi \cite{inoue2019DeploymentConscious} & - &  - & 95.70   \\
FPCnet \cite{liu2019FPCNetFast}& - &  - & 96.93  \\
\hline
Ours, no TTA & 97.36\textpm0.26  & 96.88\textpm0.32 & 96.97\textpm0.23 \\
\textbf{Ours + TTA } & \textbf{97.64\textpm0.26} & \textbf{96.91\textpm0.36} & \textbf{97.03\textpm0.30}  \\
\hline
\end{tabular}
\end{table}

\subsubsection{Results on CT260, CRKWH100, Stone331}
\label{sec:results_ct}
Only one other work \cite{zou2019DeepCrackLearning} makes use of the CT260, CRKWH100 and Stone331 datasets and reports their results. Our evaluation results are shown in \tablename \ \ref{table_ct260}. Without TTA our method performs on average 2.03\% on the \textit{OIS} and 2.61\% better on the \textit{ODS} metric on CRKWH100, achieving 93.20\% and 92.66\% respectively. Adding TTA then increases the average results by a larger margin to 94.25\% \textit{OIS} and 93.34\% \textit{ODS}. On the Stone331 evaluation dataset, without using TTA, we also report increased results, achieving 87.58\% \textit{OIS} and 87.36\% \textit{ODS}, in comparison with 87.51\% \textit{OIS} and 85.59\% \textit{ODS} of the previous best results. However, if we then enable the use of TTA on this dataset we achieve 93.30\% \textit{OIS} and 92.17\% \textit{ODS}, outperforming the previous approach by a margin of 5.79\% \textit{OIS} and 6.58\% \textit{ODS}.

\begin{table}[htbp]
\centering
\footnotesize

    \caption{Results on CRKWH100 and Stone331 when trained on CT260. Our results are reported as (Avg)\textpm(Stdev)  over 10 runs.}
    \label{table_ct260}
    
    \begin{tabular}{|l?r|r?r|r|}
        \hline
        \multicolumn{1}{|c?}{\multirow{2}{*}{\textbf{Method}}} & \multicolumn{2}{c?}{\textbf{CRKWH100}} & \multicolumn{2}{c|}{\textbf{Stone331}} \\ \cline{2-5}
         & $\boldsymbol{OIS}$ & $\boldsymbol{ODS}$& $\boldsymbol{OIS}$ & $\boldsymbol{ODS}$ \\
        \hline
        DeepCrack \cite{zou2019DeepCrackLearning} & 91.17 & 90.05 &  87.51 & 85.59  \\
        \hline
        Ours, no TTA & 93.20\textpm0.18 & 92.66\textpm0.18 & 87.58\textpm0.91 & 87.36\textpm0.83 \\
        \textbf{Ours + TTA }  & \textbf{94.25\textpm0.13}   & \textbf{93.34\textpm0.07} & \textbf{93.30\textpm0.52 } &  \textbf{92.17\textpm0.48} \\
        \hline
    \end{tabular}

\end{table}

\subsubsection{Results on DeepCrack-DB}
\label{sec:results_db}
To the best of our knowledge, only the authors of \cite{liu2019DeepCrackDeep} have used and reported their results on the DeepCrack-DB utilizing their proposed train-test split, hence we only compare against this other work. Our results on this dataset, as shown in \tablename \ \ref{table_deepcrack_db} when not using TTA indicate a slight improvement over the previous approach by 0.79\% in the \textit{cODS} metric. However, when TTA is used this margin increases to an improvement of 1.53\% in the \textit{cODS} metric. Using the \textit{OIS} and \textit{ODS} metric, we achieve 88.86\% and 85.39\% respectively. Whilst in the previous three datasets a two pixel threshold is used to count \textit{TP} pixel, in this dataset the results are reported using a zero pixel threshold, meaning a \textit{TP} pixel only counts if it lies directly on the ground truth location.

\begin{table}[htbp]
\footnotesize
\centering
    \caption{Results on DeepCrack-DB. Our results are reported as (Avg)\textpm(Stdev)  over 10 runs.}
    \label{table_deepcrack_db}
    \begin{tabular}{|l?r|r|r|}
        \hline
        \textbf{Method}  & $\boldsymbol{OIS}$ & $\boldsymbol{ODS}$& $\boldsymbol{cODS}$ \\
        \hline
        Liu \textit{et al.} \cite{liu2019DeepCrackDeep} & - &  - & 86.50 \\
        \hline
        Ours, no TTA & 87.43\textpm0.45 & 85.04\textpm0.48 & 87.29\textpm0.34   \\
        \textbf{Ours + TTA } & \textbf{88.86\textpm0.28} & \textbf{85.39\textpm0.44}& \textbf{88.03\textpm0.42}  \\
        \hline
    \end{tabular}
\end{table}

\subsection{Analysis of Different Architecture Components}
To achieve the optimal results using our proposed encoder-decoder design, we performed several side-studies, analyzing the impact of different components that make up our architecture.

Unless stated otherwise, all the studies in this section have been performed using an \textit{EfficientNet B3} encoder backbone. This provides a good trade-off in representative performance, model-size and training time and we assume that the results showing increases or decreases in performance also carry over to other backbones. 
Additionally the baseline model variations in the following studies make use of bilinear upsampling, transfer-learning, deep supervision unless it is explicitly changed.

The results are reported after using test-time-augmentation and being averaged across 10 training runs as (Avg)\textpm(Stdev), in which each training run from one to ten uses the same seed across all models. 

\subsubsection{Upsampling method}
Encoder-decoder based architectures can use multiple methods to increase the size of the feature maps in the decoder stages. Well established methods are bilinear interpolation \cite{oktay2018AttentionUNet, chen2018DeepLabSemantic}, upsampling using transposed convolutions \cite{liu2019DeepCrackDeep, long2015FullyConvolutional, konig2019SegmentationSurface} or upsampling using max-pooling indices \cite{badrinarayanan2017SegnetDeep, zou2019DeepCrackLearning}. Our analysis compared the effects of using bilinear interpolation, nearest neighbor upsampling, as well replacing the upsampling layer with a transposed convolution of kernel size $4\times4$. To not significantly increase the number of parameters in this transposed convolution, we utilize a bottleneck approach. The number of channels going in to this transposed convolution is set to one fourth of the number of input channels into the general upsampling operation and is achieved through a $1\times1$ convolution. The transposed convolution is then carried out with the number of filters matching the number of channels of its input and is followed by increasing the number of channels again through another $1\times1$ convolution, whose number of filters matches that of the input which is upsampled. Due to not all encoder architectures making use of max-pooling, comparing the use of pooling indices for upsampling has been omitted.

\tablename \ \ref{table_upsampling} shows that across all datasets the nearest neighbor and bilinear upsampling method achieve comparable performance. Upsampling using transposed convolutions performs about equally well on CRKWH100 and CFD, however on Stone331 it trails behind the former upsampling methods. It is to note that transposed convolutions incur a computational overhead as well as extra parameters to the architecture. 

Due to inducing the smallest amount of extra computation, in comparison to the other methods, as well as achieving a slightly higher performance on both metrics in two separate datasets we chose to use nearest neighbor upsampling and suggest to also use this when utilizing this method on further datasets.

\begin{table}[htbp]
\caption{Results showing the impact of using different upsampling strategies within our proposed decoder. The column header in italic highlights the method chosen in our architecture, whereas bold results indicate the highest performance in a particular category.}
\label{table_upsampling}
\footnotesize
\centering

\begin{tabular}{|l|l?r|r|r|}
\hline
\textbf{Dataset} & \textbf{Metric} & \parbox{1.7cm}{\vspace{1pt}\textit{Nearest}\\\textit{Upsampling}} & \parbox{1.7cm}{\vspace{1pt}Bilinear\\Upsampling} & \parbox{1.6cm}{Transposed\\Convolution\vspace{1pt}} \\
\hline
\multirow{2}{*}{\textbf{CT260 + CRKWH100}}      & OIS &  93.93\textpm0.10 &  93.94\textpm0.18 &  \textbf{93.97\textpm0.18}\\ \cline{2-5}
                                                & ODS &  93.06\textpm0.12 &  93.12\textpm0.18 & \textbf{93.12\textpm0.13} \\
\hline
\multirow{2}{*}{\textbf{CT260 + Stone331}}   & OIS &  89.30\textpm3.36 & \textbf{89.72\textpm2.16} &  87.98\textpm4.20\\ \cline{2-5}
                                            & ODS & 88.48\textpm3.17 & \textbf{88.80\textpm2.16} &  86.47\textpm4.73\\
\hline
\multirow{2}{*}{\textbf{CFD}}           & OIS &  \textbf{97.73\textpm0.19} & 97.69\textpm0.13 &  97.68\textpm0.15\\ \cline{2-5}
                                        & ODS & \textbf{97.07\textpm0.28} & 97.00\textpm0.19 &   97.05\textpm0.21\\
\hline
\multirow{2}{*}{\textbf{DeepCrack-DB}}  & OIS & \textbf{88.91\textpm0.16} & 88.90\textpm0.19 &  88.88\textpm0.18\\ \cline{2-5}
                                        & ODS & \textbf{85.49\textpm0.33} & 85.42\textpm0.46 &  85.36\textpm0.30\\
\hline
\end{tabular}

\end{table}

\subsubsection{Transfer Learning and Deep Supervision}
We compare training our proposed encoder-decoder architecture from scratch against utilizing pretrained weights in the encoder. In this transfer learning approach, the weights of the EfficientNet implementation that has been pretrained on the ImageNet \cite{russakovsky2015ImageNetLarge} dataset for classification, are used. In addition to that we also show the impact that deep supervision has on the performance of the model. 
As our results in \tablename \ \ref{table_ablation} indicate, using the model without deep supervision yields slightly worse performance across all datasets. When comparing using a transfer learning approach with initializing the encoder weights from scratch it is shown that pretrained encoder-weights increase the performance significantly in most of the datasets but DeepCrack-DB, where only a minor increase is achieved. We argue that this is due to the strong feature extraction baseline which backbones trained on ImageNet provide.

\begin{table}[htbp]
\caption{Results showing the impact of using deep supervision (DSV) and transfer learning (TL). The column header in italic highlights the method chosen in our architecture, whereas bold results indicate the highest performance in a particular category.}
\label{table_ablation}
\centering
\footnotesize
\begin{tabular}{|l|l?r|r|r|}
\hline
\textbf{Dataset} & \textbf{Metric} & \parbox{1.7cm}{\vspace{1pt}\textit{Ours} \\+ \textit{DSV} \\+ \textit{TL}} & \parbox{1.7cm}{Ours \\ without DSV} & \parbox{1.5cm}{Ours \\ without TL} \\
\hline
\multirow{2}{*}{\textbf{CT260 + CRKWH100}}   & OIS & \textbf{93.93\textpm0.10} & 93.78\textpm0.39 & 92.24\textpm0.88 \\ \cline{2-5}
                                    & ODS & \textbf{93.06\textpm0.12} & 92.74\textpm0.42 & 91.35\textpm1.08 \\
\hline
\multirow{2}{*}{\textbf{CT260 + STONE331}}   & OIS & \textbf{89.30\textpm3.36} & 81.76\textpm5.93 & 62.01\textpm39.74 \\ \cline{2-5}
                                    & ODS & \textbf{88.48\textpm3.17} & 80.28\textpm6.18 & 61.00\textpm39.50 \\
\hline
\multirow{2}{*}{\textbf{CFD}}                & OIS & \textbf{97.73\textpm0.19} & 97.28\textpm0.11 & 95.49\textpm0.95 \\ \cline{2-5}
                                    & ODS & \textbf{97.07\textpm0.28} & 96.49\textpm0.22 & 94.54\textpm1.14\\ 
\hline
\multirow{2}{*}{\textbf{DeepCrack-DB}}       & OIS & \textbf{88.91\textpm0.16} & 88.64\textpm0.17 & 88.34\textpm0.15 \\ \cline{2-5}
                                    & ODS & \textbf{85.49\textpm0.33} & 84.66\textpm0.36 & 84.81\textpm0.29 \\
\hline
\end{tabular}

\end{table}

\subsubsection{Choice of Feature Encoder Backbone}

The impact on performance of utilizing our proposed decoder with a variety of different encoder backbones can be seen in \tablename \ \ref{table_backbone}. Here, we use a selection of popular image classification backbones, whose weights have been pretrained on ImageNet. By comparing EfficientNet \textit{B1 - B5} backbones it is shown that using larger EfficientNet backbones increases results on the CRKWH100 and Stone331 validation datasets, whereas similar performance is achieved on CFD and DeepCrack-DB. 
When comparing the performance of the largest, \textit{EfficientNet B5} backbone with a \textit{VGG 19} or\textit{ ResNet 50} backbone, the EfficientNet based one 
generally performs better in all but, the CFD, where using a \textit{ResNet 50} encoder yields a slightly higher performance.

Our choice to report results in Tables \ref{table_cfd}, \ref{table_ct260} and \ref{table_deepcrack_db} using an \textit{EfficientNet B5} backbone is due to the large increase in performance in the CRKWH100 and Stone331 validation datasets and similar performance to the best results by other backbones in the other datasets.
Comparing the performance of models with similar number of parameters and different backbones (e.g. \textit{EfficientNet B4} and \textit{VGG 19} as well as \textit{EfficientNet B5} and \textit{ResNet 50}) it can be seen that the models with EfficientNet based backbones generally perform better in three out of the four datasets. We also report a high fluctuation in results when trained on CT260 and tested on Stone331 when using the smallest EfficientNet backbone, \textit{B0}. However, when it is tested on CRKWH100 there are more consistent results. We attribute this to overfitting the small EfficientNet backbone on the training dataset and assume that it is because the CRKWH100 test-set and the CT260 training set are of similar nature, whereas the Stone331 dataset appears different as described in section \ref{sec:datasets}.

\setlength{\tabcolsep}{4pt}
\begin{table}[h]
\caption{Results showing the impact of using different encoder-backbones with our decoder. The backbone in italic highlights the method chosen in our architecture, whereas bold results indicate the highest performance in a particular category.}
\label{table_backbone}
\centering
\footnotesize
\begin{tabular}{|l|r?r|r?r|r|}
\hline
    % Column Headers
    \multicolumn{1}{|c|}{\multirow{2}{*}{\textbf{Backbone}}} &
    \multicolumn{1}{c?}{\multirow{2}{*}{\parbox{1.3cm}{\#\textbf{Model Params}}}} &
    \multicolumn{2}{c?}{\textbf{CT260 + CRKWH100}} &
    \multicolumn{2}{c|}{\textbf{CT260 + Stone331}} \\ 
    
    \cline{3-6}& &
    $\boldsymbol{OIS}$ & $\boldsymbol{ODS}$ &
    $\boldsymbol{OIS}$ & $\boldsymbol{ODS}$ \\
\hline

EfficientNet B0 & 12.5M & 
    90.12\textpm1.49 & 88.55\textpm1.65 & %CRWKH100
    26.46\textpm22.53 & 24.64\textpm21.85 \\ %Stone331
EfficientNet B1 & 15.0M & 
    92.90\textpm1.32 & 92.03\textpm1.45 & %CRWKH100
    87.43\textpm3.44 & 86.48\textpm3.35 \\ %Stone331
EfficientNet B2 & 16.7M & 
    93.50\textpm0.28 & 92.56\textpm0.21 & %CRWKH100
    89.32\textpm2.18 & 88.53\textpm1.99 \\ %Stone331
EfficientNet B3 &  20.2M & 
    93.93\textpm0.10 & 93.06\textpm0.12 & %CRWKH100
    89.30\textpm3.36 & 88.48\textpm3.17 \\ %Stone331
VGG 19 & 26.7M & 
    93.11\textpm0.19 & 92.59\textpm0.25 & %CRWKH100
    91.56\textpm0.67 & 90.05\textpm0.94 \\ %Stone331
EfficientNet B4 & 28.1M & 
    93.70\textpm0.16 & 93.05\textpm0.21 & %CRWKH100
    91.09\textpm2.25 & 89.96\textpm3.17 \\ %Stone331
ResNet 50 & 35.0M & 
    93.60\textpm0.06 & 93.16\textpm0.13 & %CRWKH100
    86.17\textpm1.46 & 86.04\textpm1.45 \\ %Stone331
\textit{EfficientNet B5} & 39.8M & 
    \textbf{94.25\textpm0.13} & \textbf{93.34\textpm0.07} & %CRWKH100
    \textbf{93.30\textpm0.52} & \textbf{92.17\textpm0.48} \\ %Stone331
\hline
\end{tabular}

\begin{tabular}{|l|r?r|r?r|r|}

\hline
    % Column Headers
    \multicolumn{1}{|c|}{\multirow{2}{*}{\textbf{Backbone}}} &
    \multicolumn{1}{c?}{\multirow{2}{*}{\parbox{1.3cm}{\#\textbf{Model Params}}}} &
    \multicolumn{2}{c?}{\textbf{CFD}} &
    \multicolumn{2}{c|}{\textbf{DeepCrack-DB}} \\ 
    
    \cline{3-6}& &
    $\boldsymbol{OIS}$ & $\boldsymbol{ODS}$ &
    $\boldsymbol{OIS}$ & $\boldsymbol{ODS}$ \\
\hline

EfficientNet B0 & 12.5M & 
    97.64\textpm0.10 & 96.98\textpm0.21 & %CFD
    88.88\textpm0.18 &  85.51\textpm0.50\\   %DeepcrackDB
EfficientNet B1 & 15.0M & 
    97.60\textpm0.16 & 96.90\textpm0.15 & %CFD
    89.03\textpm0.17 & \textbf{85.61\textpm0.31}  \\   %DeepcrackDB
EfficientNet B2 & 16.7M & 
    97.75\textpm0.07 & 97.22\textpm0.15 & %CFD
    89.02\textpm0.17 & 85.34\textpm0.40 \\   %DeepcrackDB
EfficientNet B3 &  20.2M & 
    97.73\textpm0.19 & 97.07\textpm0.28 &    %CFD
    88.91\textpm0.16 & 85.49\textpm0.33  \\   %DeepcrackDB
VGG 19 & 26.7M & 
    96.53\textpm0.29 & 95.75\textpm0.31 & %CFD
    88.10\textpm0.15 &  84.65\textpm0.22 \\   %DeepcrackDB    
EfficientNet B4 & 28.1M & 
    97.86\textpm0.10 & 97.17\textpm0.14 & %CFD
    \textbf{89.05\textpm0.17} & 85.49\textpm0.38  \\   %DeepcrackDB
ResNet 50 & 35.0M & 
    \textbf{97.92\textpm0.09} & \textbf{97.37\textpm0.18} & %CFD
    88.68\textpm0.17 & 84.73\textpm0.39  \\   %DeepcrackDB    
\textit{EfficientNet B5} & 39.8M & 
    97.64\textpm0.26 & 96.91\textpm0.36 & %CFD
   88.86\textpm0.28 & 85.39\textpm0.44 \\   %DeepcrackDB
  \hline
\end{tabular}

\end{table}

\subsection{Impact of Test-Time-Augmentation}
As shown in Sections \ref{sec:results_cfd}, \ref{sec:results_ct} and \ref{sec:results_db} the application of TTA leads to a mixed performance improvement ranging from 0.24\% OIS on CFD to 5.72\% OIS on Stone331. We hypothesize that the difference in dataset appearances as well as crack sizes may be causing this variation. Whereas CFD has a high diversity in crack sizes, a majority of the cracks in Stone331 are only a single pixel wide. It is assumed that TTA facilitates a better segmentation of very thin cracks as enlarging them may make them easier to detect for the algorithm.

\subsection{Generalization Abilities}

The results in the previous sections show, that the model achieves high performance on all datasets tested. Additionally, to gain further insight into the model performance, we have also studied the generalization abilities of our proposed encoder-decoder model. 
We show the results in \tablename \ \ref{table_generalizattion}. Here, we used the models which have been trained on one specific dataset and evaluate them on all other testing-datasets. 

Overall our model achieves high generalization performance on all but three configurations: 
Training on CFD or DeepCrack-DB  and testing on Stone331 leads to a low performance and training on CT260 and testing on DeepCrack-DB also leads to a low generalization performance in both metrics. 
For the first two configurations we attribute this low performance on the difference in background textures inbetween the CFD/DeepCrack-DB and Stone331 datasets. For the third configuration we assume that the crack widths in CT260, with a large majority of cracks only being a single pixel wide, differs too much from the data appearing in DeepCrack-DB, where the crack width can span up to several dozen pixels. 

Therefore, to achieve the best possible performance using this model, the training data should include cracks of different widths, ranging from one- to several pixels width, and a large diversity in backgrounds textures.

\begin{table}[htbp]
\caption{Generalization results inbetween different datatsets. Cells with gray background indicate the original train-test dataset split, whereas cells with a white background highlight the generalization results.}
\label{table_generalizattion}
\centering
\footnotesize

\begin{tabular}{|l|r|r?r|r|}
\hline
    % Column Headers
    \multicolumn{1}{|l|}{\multirow{2}{*}{\diagbox[innerwidth=2.5cm]{Train}{Test}}} &
    \multicolumn{2}{c?}{\textbf{CRKWH100}} &
    \multicolumn{2}{P{3.5cm}|}{\textbf{Stone331}} \\ 
    
    \cline{2-5}& 
    $\boldsymbol{OIS}$ & $\boldsymbol{ODS}$ &
    $\boldsymbol{OIS}$ & $\boldsymbol{ODS}$ \\
\hline

\textbf{CT260} & 
    \cellcolor{gray} 94.25\textpm0.13 & \cellcolor{gray} 93.34\textpm0.07 & %CRWKH100
    \cellcolor{gray} 93.30\textpm0.52 & \cellcolor{gray} 92.17\textpm0.48 \\ %Stone331
\hline
\textbf{CFD} & 
    94.02\textpm0.80 & 89.08\textpm1.55 & %CRWKH100
    36.79\textpm11.42 & 28.13\textpm10.35 \\ %Stone331
\hline
\textbf{DeepCrack-DB} & 
    89.37\textpm1.25 & 85.44\textpm1.28 & %CRWKH100
    15.97\textpm2.98 & 10.98\textpm1.83 \\ %Stone331
\hline
\end{tabular}

\begin{tabular}{|l|r|r?r|r|}
\hline
    % Column Headers
    \multicolumn{1}{|l|}{\multirow{2}{*}{\diagbox[innerwidth=2.5cm]{Train}{Test}}} &
    \multicolumn{2}{c?}{\textbf{CFD}} &
    \multicolumn{2}{P{3.5cm}|}{\textbf{DeepCrack-DB}} \\ 
    
    \cline{2-5}& 
    $\boldsymbol{OIS}$ & $\boldsymbol{ODS}$ &
    $\boldsymbol{OIS}$ & $\boldsymbol{ODS}$ \\
\hline

\textbf{CT260} & 
    96.12\textpm0.56 &  96.04\textpm0.58 & %CFD
    28.86\textpm0.82 &  28.86\textpm0.82 \\ %DeepcrackDB
\hline
\textbf{CFD} & 
    \cellcolor{gray} 97.64\textpm0.26 & \cellcolor{gray} 96.91\textpm0.36 & %CFD
    79.70\textpm3.44 & 75.50\textpm3.49 \\ %DeepcrackDB
\hline
\textbf{DeepCrack-DB} & 
    94.02\textpm0.78 & 93.81\textpm0.75 & %CFD
    \cellcolor{gray} 88.86\textpm0.28 & \cellcolor{gray} 85.39\textpm0.44 \\ %DeepcrackDB
\hline
\end{tabular}
\end{table}

\subsection{Segmentation Faults}

\begin{figure}[!ht]
\centering
\includegraphics[width=0.6\linewidth]{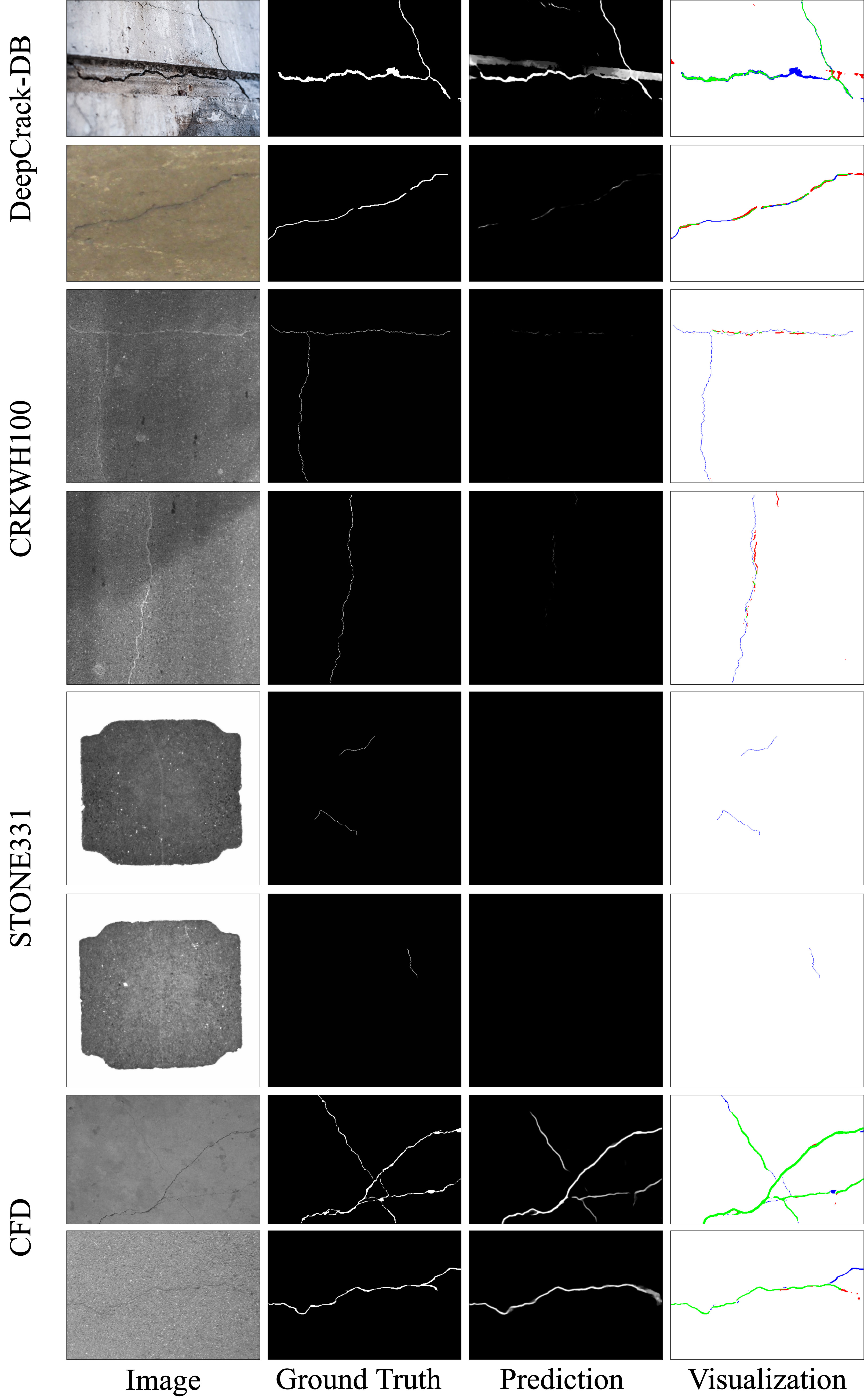}
\caption{Sample images where our proposed model does not generate adequate segmentation results. Two images from each dataset (DeepCrack-DB, CRKWH100, Stone331 and CFD, from top to bottom) are shown with their respective ground truth, a sample prediction of our proposed method and a results visualization at the OIS. Here, green pixels are $TP$, blue pixels $FN$ and red pixels $FP$. Note that in the CFD, CRKWH100 and Stone331 samples we use the pixel distance threshold of 2, meaning predictions are $TP$ pixels if they lie within 2 pixels to a ground truth, $FP$ are incorrect predictions more than 2 pixel away from a ground truth and $FN$ are missed ground truth pixels if they are further than 2 pixels away from a predicted pixel. (Best viewed digitally and in color).}
\label{fig_failures}
\end{figure}

To outline the limitations of our proposed method we have included samples where our algorithm failed to generate adequate segmentation maps. These are shown in \figurename \ \ref{fig_failures}. Samples were selected as they have the the worst OIS scores (averaged over all 10 runs) per dataset and give some insight into where our algorithm fails to segment. 

As it can be seen on the first DeepCrack-DB sample, the algorithm has difficulties distinguishing very dark linear areas from cracks. Additionally, the second sample shows low performance on an image where the crack has very low contrast compared to the background. In the following two CRKWH100 samples, this method achieves a low performance due to the cracks being bright, which is in contrast to the majority of crack samples in the training set appearing darker. The Stone331 samples show no crack-detection at all, leading to an $OIS$ score of 0 on those images. We argue that is because the cracks appear very similar to their background. This problem also appears in the CFD samples, where crack regions which are very similar to their background are barely detected. 

Generally, the limitations of this method lie in the detection of cracks that appear very different from the ones that appear in the training dataset, as well ones which have a low contrast between the background and the cracks. Both of those limitations can be mitigating by providing more diverse training data, as exposure to more different samples will lead to better testing-performance.

\color{black}

\subsection{Model Efficiency and Complexity}

When utilizing an \textit{EfficientNet B5} backbone our model has a total of 39.8 Million parameters, with the decoder consisting of 11.3 million parameters and the backbone containing 28.5 million parameters. The number of FLOPS (multiply-adds) of this model is 42.9 billion for a single channel grayscale image of size $512\times512$ pixel.
During inference, without using test-time-augmentation, on our hardware, the model architecture can create predictions with a speed of 11.7 frames per second on images of the previously mentioned size. When utilizing test-time-augmentation, the inference speeds drops to 1.9 frames per second, due to creating and merging of multiple segmentation maps per image. Real-time applications might therefore not benefit from that approach and it seems to be better suited for batch-processing.
The approximate training time of our proposed method using an \textit{EfficientNet B5} backbone for a single run with the aforementioned configuration is 1:20h for CFD, 4:50h for CT260 and 5:30h for Deepcrack-DB.

\section{Conclusion}
\label{sec:conclusion}
In this work we propose novel surface crack segmentation methods using an encoder-decoder based deep learning architecture. We introduce a decoder design 
which can be added to existing feature encoder backbone architectures such as ResNet, VGG or EfficientNet to complete the U-shape of the network. We also propose the use of test-time-augmentation and performing a statistical analysis over multiple training runs. Whilst both techniques may previously have been used in other fields, this is the first application in the crack-segmentation domain. It is shown that the use of test-time-augmentation generally increases the models performance and performing this statistical analysis allows for better reproducibility, as it captures a models performance better rather than reporting only one results, which is common in this field \cite{zou2019DeepCrackLearning, liu2019FPCNetFast, liu2019DeepCrackDeep, inoue2019DeploymentConscious}.
We show that our optimized deep encoder-decoder methods using a pretrained \textit{EfficientNet B5} backbone as an encoder with our proposed decoder, new state of the art results in four different crack segmentation datasets are achieved.  Namely we outperform the previous best approaches: FPCNet \cite{liu2019FPCNetFast} on the CFD \cite{shi2016AutomaticRoad} dataset, DeepCrack \cite{zou2019DeepCrackLearning} on the CRKWH100 and Stone331 datasets \cite{zou2019DeepCrackLearning} and another DeepCrack \cite{liu2019DeepCrackDeep} on DeepCrack-DB \cite{liu2019DeepCrackDeep}. By presenting our results on multiple datasets we show that our approach is well suited for crack segmentation. Additionally we also highlight the generalization abilities by cross-testing inbetween all datasets and outline the limitations which align with the results of the generalization-study: Our methods performance is limited when the appearance of cracks and their backgrounds in the training- and testing data is very different and advise for better performance to use more diverse datasets.

Further experiments show that that EfficientNet based backbones of similar size to either \textit{ResNet 50} and \textit{VGG 19} perform better on a majority of the datasets, exhibiting a better generalization performance. We also show that adding transfer learning  with pretrained ImageNet weights and using deep supervision leads to an additional performance improvement. 

In future work, we aim to study the performance of using this method in other domains as well as examine the ability to denoise crack images. We also aim to overcome the limitations of this approach by generating a much larger dataset and train this method on it. Additionally, further research may also study the direct impact of TTA and the underlying reason of different performance improvements in the various datasets.

\section*{Acknowledgments}
This work was supported by The Data Lab Industrial PhD programme.



%\section*{References}
\bibliographystyle{elsarticle-num} 
\bibliography{main}


%% \begin{thebibliography}{00}


%% \bibitem{}

%% \end{thebibliography}
\end{document}